\documentclass[runningheads]{llncs}

\usepackage{hyperref}
\usepackage[utf8]{inputenc}
\usepackage{enumitem}
\usepackage{booktabs}

\usepackage{amsmath}
\usepackage{graphicx}
\usepackage{amssymb}

\usepackage{algorithm}
\usepackage{algorithmic}

\newcommand{\mat}[1]{\boldsymbol{\mathrm{#1}}}
\newcommand{\vecb}[1]{\boldsymbol{#1}}

\newcommand{\indep}{\perp \!\!\! \perp}
\newcommand{\given}{\mid}
\newcommand{\st}{\ :\ }

\DeclareMathOperator{\pa}{pa}

\DeclareMathOperator{\nei}{ne}

\title{Bayesian optimization of the PC algorithm for learning Gaussian Bayesian
networks}
\author{Irene Córdoba\inst{1} \and Eduardo C. Garrido-Merchán\inst{2} \and Daniel
Hernández-Lobato\inst{2} \and Concha Bielza\inst{1} \and Pedro Larrañaga\inst{1}}

\authorrunning{I. Córdoba et al.}
\titlerunning{Bayesian Optimization of the PC algorithm}

\institute{Universidad Politécnica de Madrid, Departamento de
Inteligencia Artificial\and Universidad Autónoma de Madrid,
Departamento de Ingeniería Informática}

\begin{document}

\maketitle

\begin{abstract}
The PC algorithm is a popular method for learning the structure of Gaussian Bayesian 
networks. It carries out statistical tests  to determine 
absent edges in the network. It is hence governed by two parameters: (i) The type
of test, and (ii) its significance level. These parameters are usually set to 
values recommended by an expert. Nevertheless, such an approach can suffer from human 
bias, leading to suboptimal reconstruction results. In this paper we consider a more 
principled approach for choosing these parameters in an automatic way.
For this we optimize a reconstruction score evaluated on a set of different 
Gaussian Bayesian networks. This objective is expensive to evaluate and lacks a 
closed-form expression, which means that Bayesian optimization (BO) is
a natural choice. BO methods use a model to guide the search and are hence able 
to exploit smoothness properties of the objective surface. We show that the parameters 
found by a BO method outperform those found by a random search strategy and the expert 
recommendation. Importantly, we have found that an often overlooked statistical 
test provides the best over-all reconstruction results. 
\end{abstract}

\section{Introduction}\label{sec:intro}
Graphical models serve as a compact representation of the relationships between
variables in a domain. An important subclass is the Bayesian network, where
conditional independences are encoded by missing edges in a directed graph with
no cycles.  By exploiting these independences, Bayesian networks yield a modular
factorization of the joint probability distribution underlying the data. Of
particular interest are Gaussian Bayesian networks for modelling variables in a
continuous domain, which have been widely applied in real scenarios such as gene
network discovery \cite{ness2016} and  neuroscience \cite{bielza2014}. 

When learning graphical models from data, two main tasks are usually
differentiated: structure and parameter learning. The former consists in 
recovering the graph structure, and the latter amounts to
fitting the numerical quantities in the model. In Gaussian Bayesian networks, parameter
learning involves using standard linear regression theory, whereas structure
learning is not an easy task in general, given the combinatorial search space of
acyclic digraphs. There are two main approaches one can find in the literature
for structure discovery in Bayesian networks: score-and-search heuristics, where
the search space is explored looking for the network which optimizes a given score
function, and constraint-based approaches, where statistical tests are performed
in order to include or exclude dependencies between variables. 

A popular constraint-based method with consistency guarantees is the PC algorithm 
\cite{kalisch2007,spirtes2000}.  In this method, a backward stepwise testing procedure is performed for
determining absent edges in the resulting graph.  Thus, of critical importance
are the choice of the statistical test to be performed, and
the significance level at which the potential edges are going to
be tested. However, both are usually fixed after a grid search or directly set 
by expert knowledge \cite{colombo2014,kalisch2007}. In the literature on Bayesian 
network structure learning some empirical studies explore exact structure recovery 
\cite{malone2018}, the behavior of score-and-search algorithms \cite{malone2015}, and 
the impact of the significance level in the PC algorithm for high dimensional sparse
scenarios \cite{colombo2014,kalisch2007}. We are not aware, however,  of 
any research work using elaborated methods for hyper-parameters selection in this context.  

In this paper we show that Bayesian optimization (BO) can be used as an
alternative methodology for choosing the significance level and the statistical 
test in the PC algorithm. BO has been recently applied
successfully in different
optimization problems \cite{shahriari2016,snoek2012practical}. We consider here 
a structure learning scenario in moderately sparse settings that is 
representative of those considered in \cite{kalisch2007}. We
show that BO outperforms, in terms of structure recovery error, in a relatively small number of iterations, 
both a baseline approach based on a grid search and specific values set by expert knowledge 
obtained from previous results on this problem \cite{kalisch2007}. Furthermore, we also analyze what values for the statistical test and the 
significance level are recommended by the BO approach, and compare them with those often used by 
the relevant literature on the subject. 

This article is organized as follows. In Section \ref{sec:prel}, we introduce
the main concepts relative to Gaussian Bayesian networks that will be
used throughout the rest of the paper. Then, in Section \ref{sec:pc}, we
describe the PC algorithm, emphasizing its hyper-parameters and how they may
affect its performance. Black box BO is outlined
in Section \ref{sec:bo}, with emphasis on the particular
characteristics of our problem. The experimental setting as well as the
results we have obtained are described in Section \ref{sec:exp}.  Finally, we
conclude the paper in Section \ref{sec:conc}, where we also point out the main
planned lines of future work.

\section{Preliminaries on Gaussian Bayesian networks}\label{sec:prel}
Throughout the remainder of the paper,
$X_1, \ldots, X_p$ will denote $p$ random variables, and $\vecb{X}$ the random
vector they form. For a subset of indices $I \subseteq \{1, \ldots, p\}$,
$\vecb{X}_I$ will denote the random vector corresponding only to the variables
indexed by $I$. We will use
$\vecb{X}_I \indep \vecb{X}_J \given \vecb{X}_K$ for denoting that $\vecb{X}_I$
is conditionally independent of $\vecb{X}_J$ given the values of $\vecb{X}_K$,
being $I, K, J$ disjoint subsets of $\{1, \ldots, p\}$.
Let $G = (V, E)$ be an acyclic digraph, where $V = \{1, \ldots, p\}$ is the
vertex set and $E \subseteq V \times V$ is the edge set. When $G$ is part of a
graphical model, its vertex set $V$ can be thought of as indexing a random
vector $\vecb{X} = (X_1, \ldots, X_p)$.  In a Bayesian network, the graph $G$ is
constrained to be acyclic directed and with no multiple edges. 

A common interpretation of edges in a Bayesian network is the ordered Markov
property, although many more exist, which can be shown to be equivalent
\cite{lauritzen1990}. This property is stated as follows. For a vertex $i \in
V$, the set of parents of $i$ is defined as $\pa(i) := \{j \st (j, i) \in E\}$.
In every acyclic digraph, an ancestral order $\prec$ can be found between the
nodes where it is satisfied that if $j \in \pa(i)$, then $j \prec i$, that is,
the parents of a vertex come before it in the ancestral order. For notational
simplicity, in the remainder we will assume that the vertex set $V = \{1,
\ldots, p\}$ is already ancestrally ordered. The ordered Markov property of
Bayesian networks can be written in this context as
\[
	X_i \indep \vecb{X}_{\{1, \ldots, i - 1\} \setminus \pa(i)} \given
	\vecb{X}_{\pa(i)}
\]
for all $i \in V$. 

The above conditional independences, together with the properties of the
multivariate Gaussian distribution, allow to express a Gaussian Bayesian network
as a system of recursive linear regressions. Indeed, if for each $i \in V = \{1, \ldots, p\}$,  
we consider the regression of $X_i$ on its predecessors
in the ancestral order, $X_1, \ldots, X_{i - 1}$, then from the results
regarding conditioning on multivariate Gaussian random variables we obtain
\begin{equation}\label{eq:linreg}
	X_i = \sum_{j = 1}^{i - 1} \beta^{ji \given 1, \ldots, i -
	1} X_j + \epsilon_i,
\end{equation}
where the regression coefficient $\beta^{ji \given 1, \ldots, i - 1} = 0$ when
$j \notin \pa(i)$, and $\epsilon_i$ are independent Gaussian random variables with 
zero mean and variance equal to the conditional variance of $X_i$ on $X_1,
\ldots, X_{i - 1}$. Therefore, both the structure and parameters of a Gaussian
Bayesian network can be directly read off from the system of linear regressions
in Equation \eqref{eq:linreg}.

\section{Structure learning with the PC algorithm}\label{sec:pc}
The PC algorithm for learning Gaussian Bayesian networks
proceeds by first estimating the skeleton, that is, the
underlying undirected graph, of the acyclic digraph, and then
orienting it. That is, for each vertex $i \in V = \{1, \ldots, p\}$, it looks
through the set of its neighbors, which we will denote as $\nei(i)$, and
selects a node $j \in \nei(i)$ and subset $C \subseteq \nei(i)\setminus\{j\}$.
Then, the conditional independence $X_i \indep X_j \given \vecb{X}_C$ is tested
on the available data. It is a backward stepwise elimination method,
in the sense that it starts with the complete undirected
graph, and then proceeds by testing conditional independences
in order to remove edges, doing so incrementally in the size of the neighbor
subset $C$. 

The PC main phase pseudocode can be found in Algorithm
\ref{alg:pcpop}. The output of Algorithm \ref{alg:pcpop} is
the skeleton, or undirected version, of the estimated Gaussian
Bayesian network, which is later oriented.
\begin{algorithm}
	\caption{The PC algorithm in its population version}
	\label{alg:pcpop}
	\begin{algorithmic}[1]
		\REQUIRE Conditional independence information about $\vecb{X} = (X_1, \ldots, X_p)$
		\ENSURE Skeleton of the Gaussian Bayesian network
		
		\STATE $G \gets$ complete undirected graph on $\{1, \ldots, p\}$
		\STATE $l \gets -1$
		\REPEAT
			\STATE $l \gets l + 1$
			\REPEAT
				\STATE Select $i$ such that $(i, j) \in E$ and $\lvert
				\nei(i)\setminus\{j\}\rvert \geq l$
				\REPEAT
					\STATE Choose new $C \subseteq \nei(i) \setminus \{j\}$ with
					$\lvert C \rvert = l$
					\IF {$X_i \indep X_j \given \vecb{X}_C$} 
						\STATE $E \gets E \setminus \{(i, j), (j, i)\}$
					\ENDIF
				\UNTIL{$(i,j)$ has been deleted or all neighbor subsets of size $l$ have been tested}
			\UNTIL{All $(i, j) \in E$ such that $\lvert
			\nei(i)\setminus\{j\}\rvert \geq l$ have been tested}
		\UNTIL{$\lvert \nei(i)\setminus\{j\}\rvert < l$ for all $(i, j) \in E$}
	\end{algorithmic}
\end{algorithm}
Algorithm \ref{alg:pcpop} is typically called the \emph{population} version of
the PC algorithm \cite{kalisch2007}, since it assumes that perfect information is
available about the conditional independence relationships present in the
data. This is useful for illustrating the behavior and main properties of the
algorithm; however, in real scenarios this is unrealistic, and statistical tests
must be performed on the data in order to determine which variable pairs, with
respect to different node subsets, are conditionally independent.

\subsection{Significance level and statistical test}
The criteria for removing edges is related to the ordered Markov property and
Equation \eqref{eq:linreg}. In particular, from multivariate Gaussian analysis
we know that for $i \in V$ and $j < i$, 
\[
	\beta^{ji \given 1, \ldots, i - 1} = 0 \iff \rho^{ji
	\given 1, \ldots, i - 1} = 0,
\]
where $\rho^{ji \given 1, \ldots, i - 1}$ denotes the partial correlation
coefficient between $X_i$ and $X_j$ with respect to $X_1, \ldots, X_{i - 1}$. In
the PC algorithm, at iteration $l$, the null hypothesis $H_0 : \rho^{ji \given
C} = 0$ is tested against the alternative hypothesis $H_1 : \rho^{ji \given C}
\neq 0$, where $C$ is a subset of the neighbors of $i$
(excluding $j$) in the current estimator of the skeleton such that $\lvert C \rvert = l$.

The significance level at which $H_0$ will be tested, which we will denote in
the remainder as $\alpha$, is typically smaller or equal than $0.05$,
and serves to control the type I error. The other parameter of
importance is the statistical test itself. The usual choice for this is a Gaussian test based on 
the Fisher's Z transform of the partial correlation coeficient \cite{colombo2014,kalisch2007},
which is asymptotically normal.
However, there are other choices available in the literature that could be
considered and can be found in standard implementations of the algorithm. For
example, the \texttt{bnlearn} R package \cite{scutari2010}
provides the standard Student's t test for the untransformed partial correlation coeficient, and the
$\chi^2$ test  and a test based on the shrinkage 
James-Stein estimator, for the mutual information \cite{hausser2009}.

\subsection{Evaluating the quality of the learned structure}
When performing structure discovery in graphical models, there are several ways
of evaluating the results obtained by an algorithm. As a starting point, one
could use standard error rates, such as the true positive and false positive
rates.  These rates simply take into account the original acyclic digraph $G =
(V, E)$, and the estimated one $\hat{G}$, with edge set $\hat{E}$. Then, 
$E$ with $\hat{E}$ are compared element-wise. This is a common approach in Bayesian networks.

We have preferred however to use the Structural Hamming Distance
(SHD) \cite{tsamardinos2006}. This measure is motivated as follows.  In Bayesian
networks, there is not a unique correspondence between the model and the acyclic
digraph that represents it.  That is, if we denote as $\mathcal{M}(G)$ the set
of multivariate Gaussian distributions whose conditional independence model is
compatible (in the sense of the pairwise Markov property and Equation
\eqref{eq:linreg}) with the acyclic digraph $G$, then we may have two distinct
acyclic digraphs $G_1$ and $G_2$ such that $\mathcal{M}(G_1) =
\mathcal{M}(G_2)$. In such case, $G_1$ and $G_2$ are said to be Markov
equivalent. 

The SHD measure between two acyclic digraph structures $G_1$ and $G_2$ takes
into account this issue of non unique correspondence. In
particular, it counts the number of operations that have to be
performed in order to transform the Markov equivalence class
of one graph into the other. Thus, given two
acyclic digraphs that are distinct but Markov equivalent, their true
positive and false positive rates could be nonzero, while their SHD is 
guaranteed to be zero.

\section{Black-box Bayesian optimization}\label{sec:bo}
{Denote the SHD objective function as $f(\vecb{\theta})$,
which depends of the parameters in the PC algorithm, $\vecb{\theta}
= (\alpha, T)$, that are
going to be optimized, $\alpha$, the significance level, and
$T$, the independence test.}
{We can view $f(\vecb{\theta})$ as a} black-box objective
{function} with noisy
evaluations $y_i = f(\vecb{\theta})+\epsilon_i$, with $\epsilon_i$
being a, typically, Gaussian noise term. {With} BO
the number of evaluations of $f$ needed to solve the
optimization problem {are drastically reduced}. Let the
observed data until step $t-1$ of the algorithm be
{$\mathcal{D}_{t-1}=\{(\vecb{\theta}_i,
y_i)\}_{i=1}^{t-1}$}. At iteration $t$ of BO, a probabilistic
model {$p(f(\vecb{\theta}) \given \mathcal{D}_{t-1})$}, typically a Gaussian process
(GP) \cite{rasmussen2004gaussian}, is fitted to the data collected
so far. The uncertainty about $f$ provided by
the probabilistic model is then
used to generate an acquisition function $a_t(\vecb{\theta})$, whose value at each
input location indicates the expected utility of evaluating
$f$ there. Therefore, at iteration $t$,
$\vecb{\theta}_t$ is chosen as the one that maximizes the acquisition
function.
The described process is
repeated until enough data about the objective has been collected. When this is
the case, the GP predictive mean for $f(\cdot)$ can be optimized to find the
solution of the optimization problem, or we can provide as a solution the best
observation made so far. 

\begin{figure}[htb!]
\centering
\begin{tabular}{c}
\includegraphics[width=0.3\textwidth]{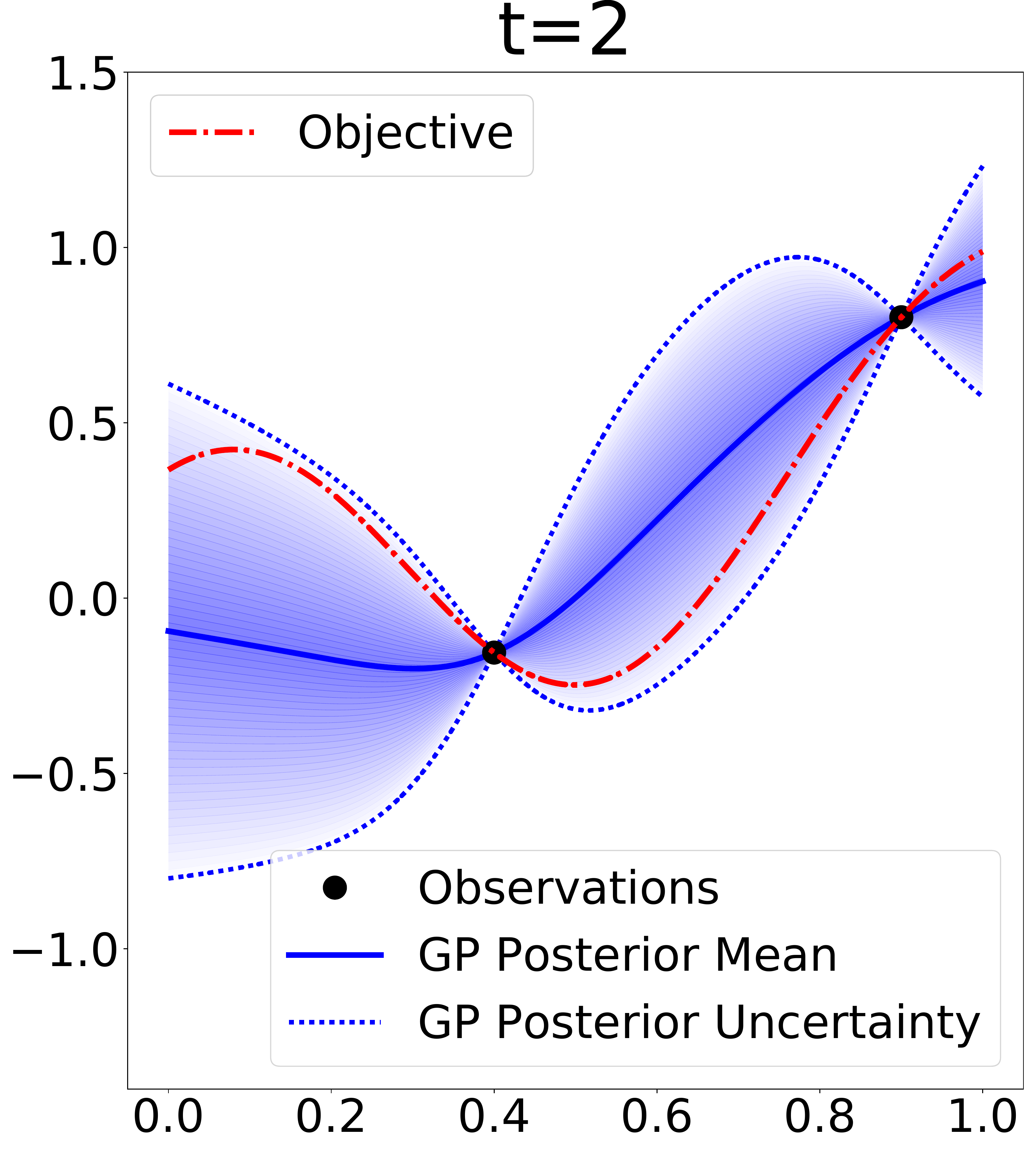} 
\includegraphics[width=0.3\textwidth]{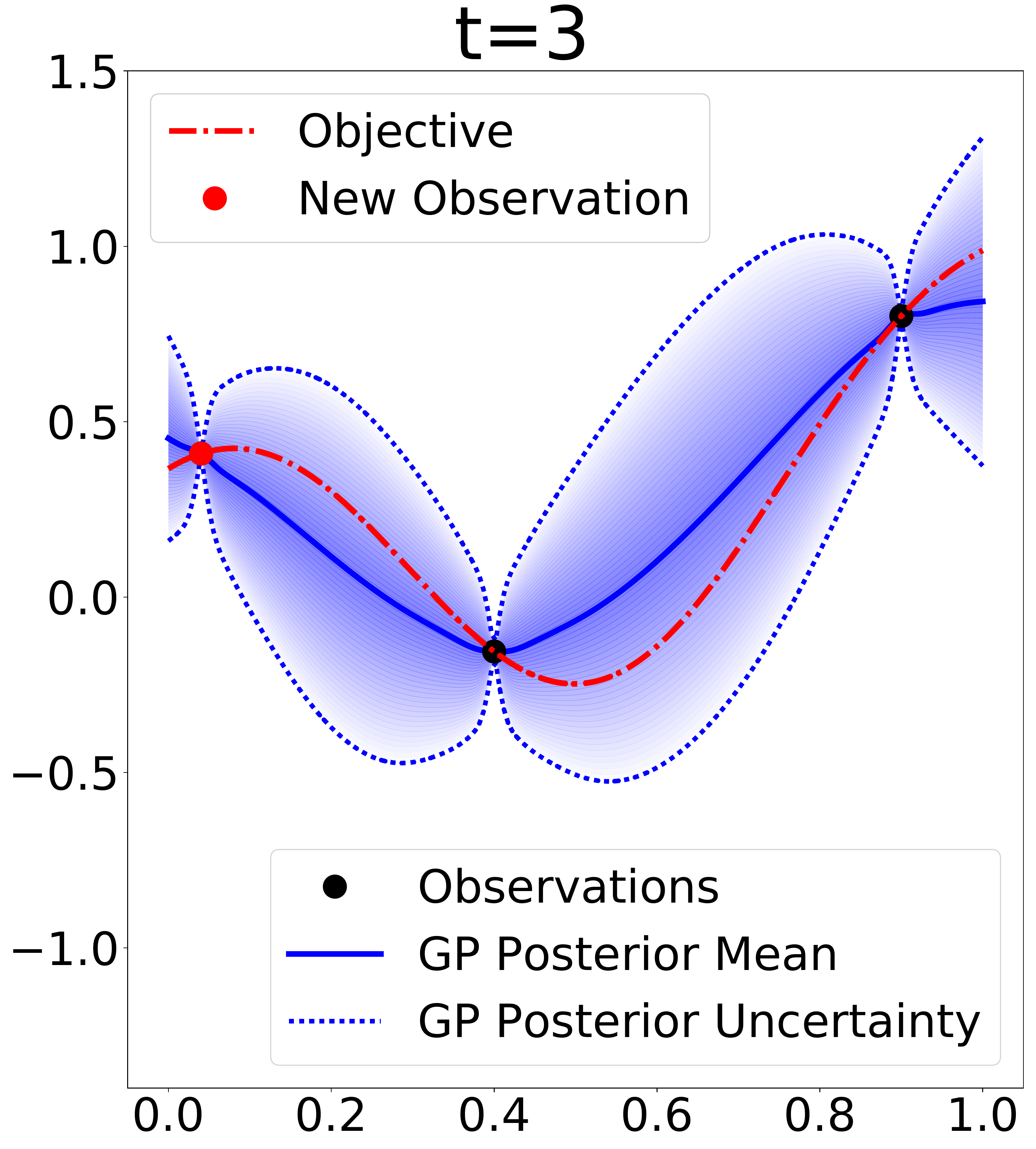} 
\includegraphics[width=0.3\textwidth]{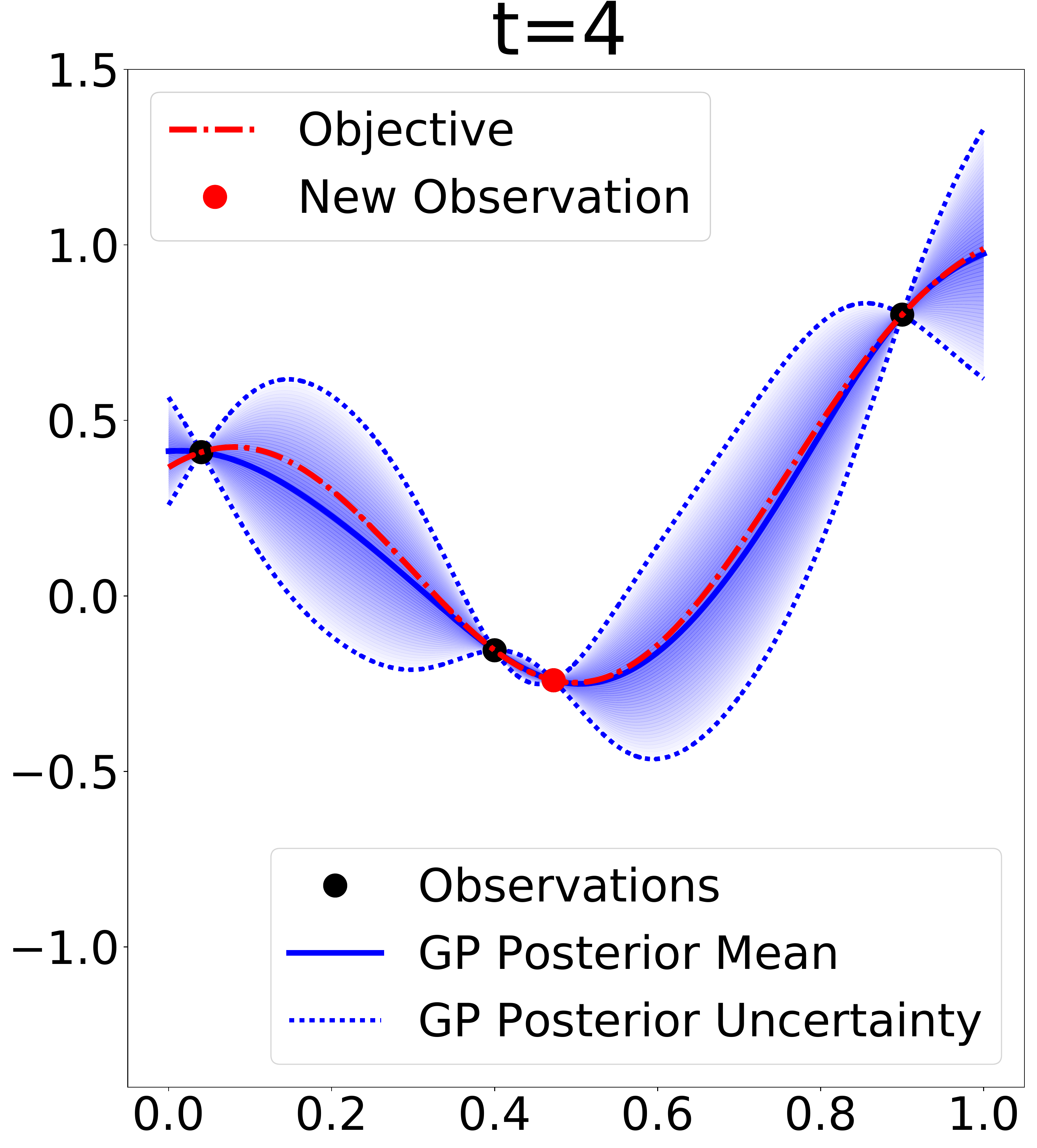} \\
\includegraphics[width=0.3\textwidth]{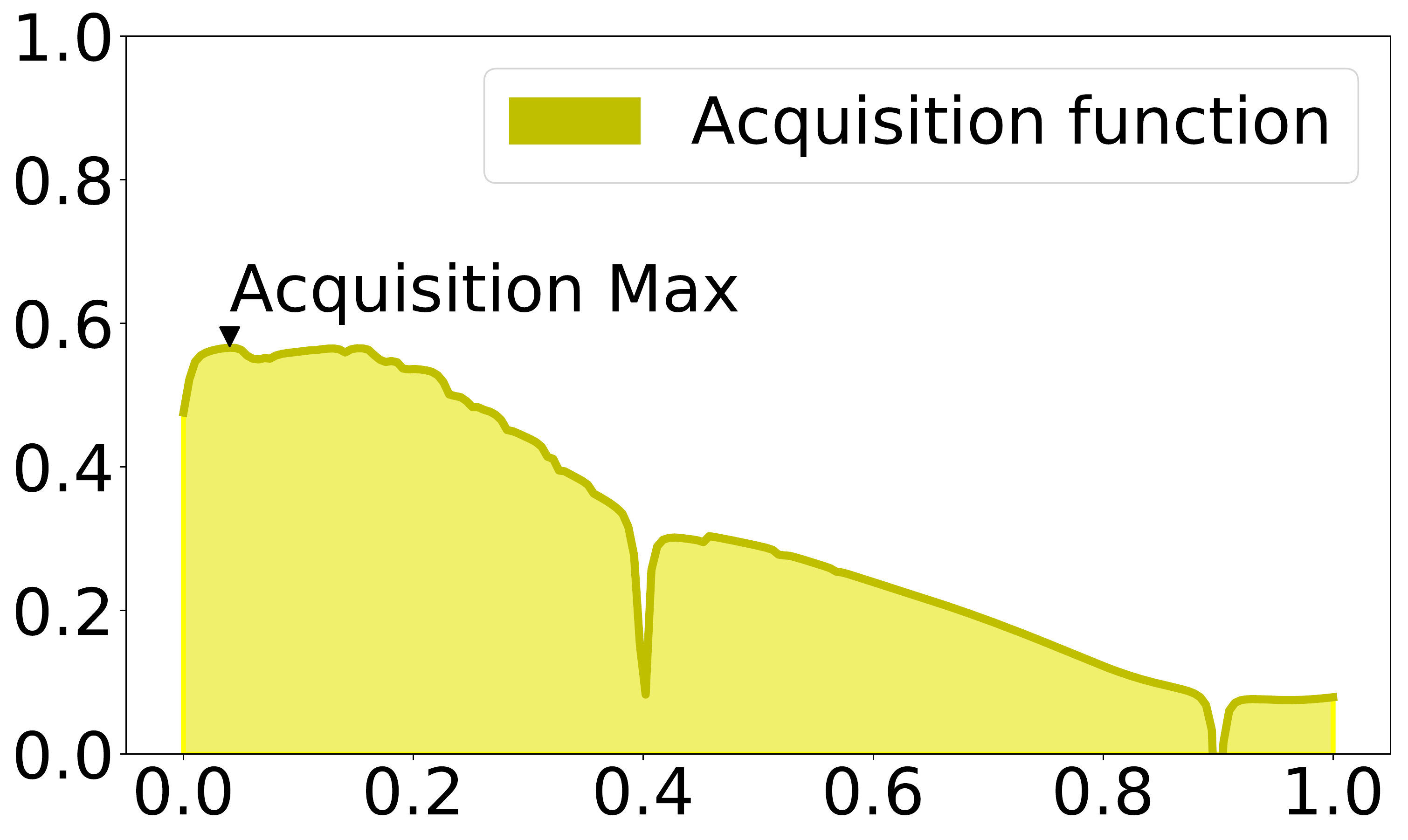} 
\includegraphics[width=0.3\textwidth]{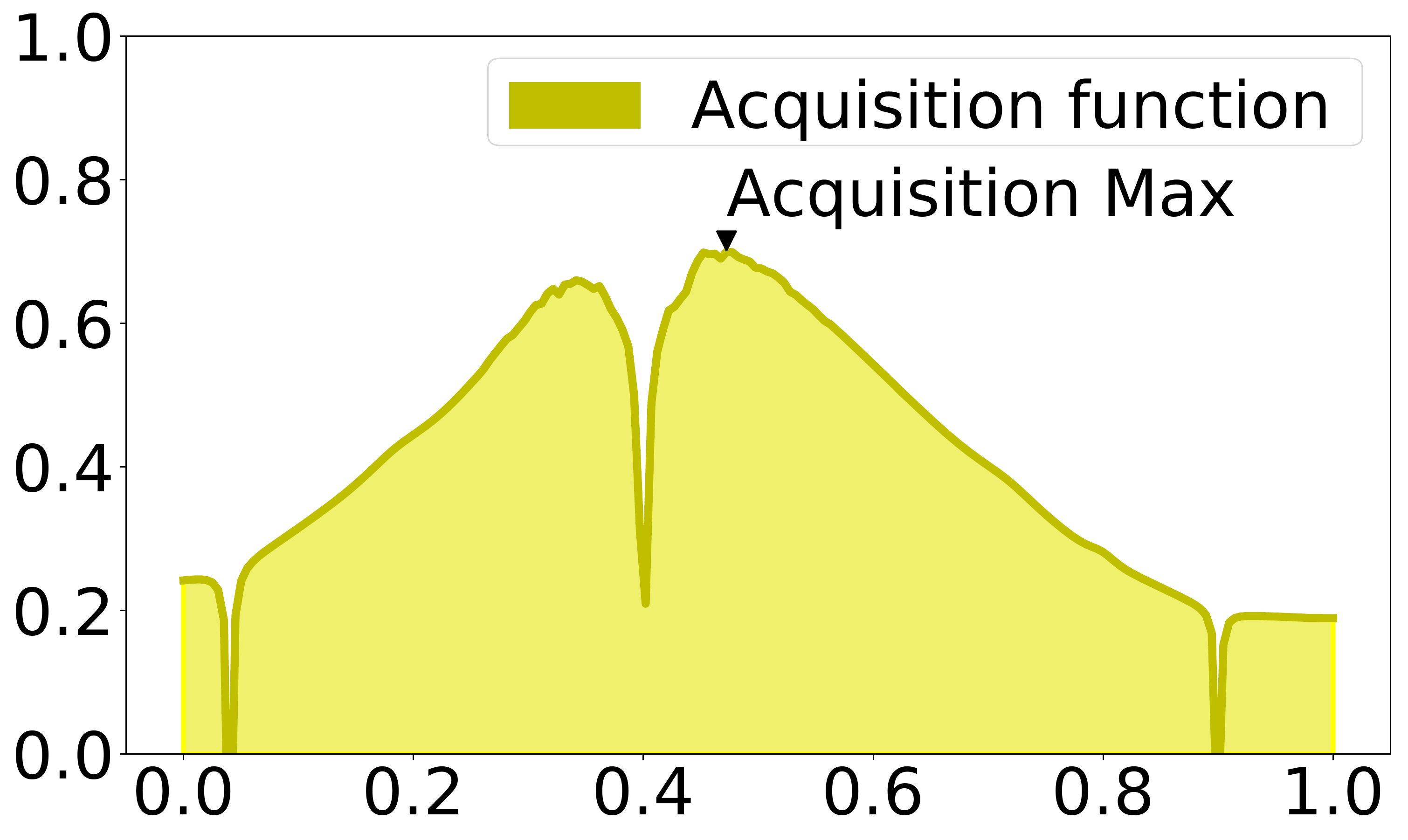} 
\includegraphics[width=0.3\textwidth]{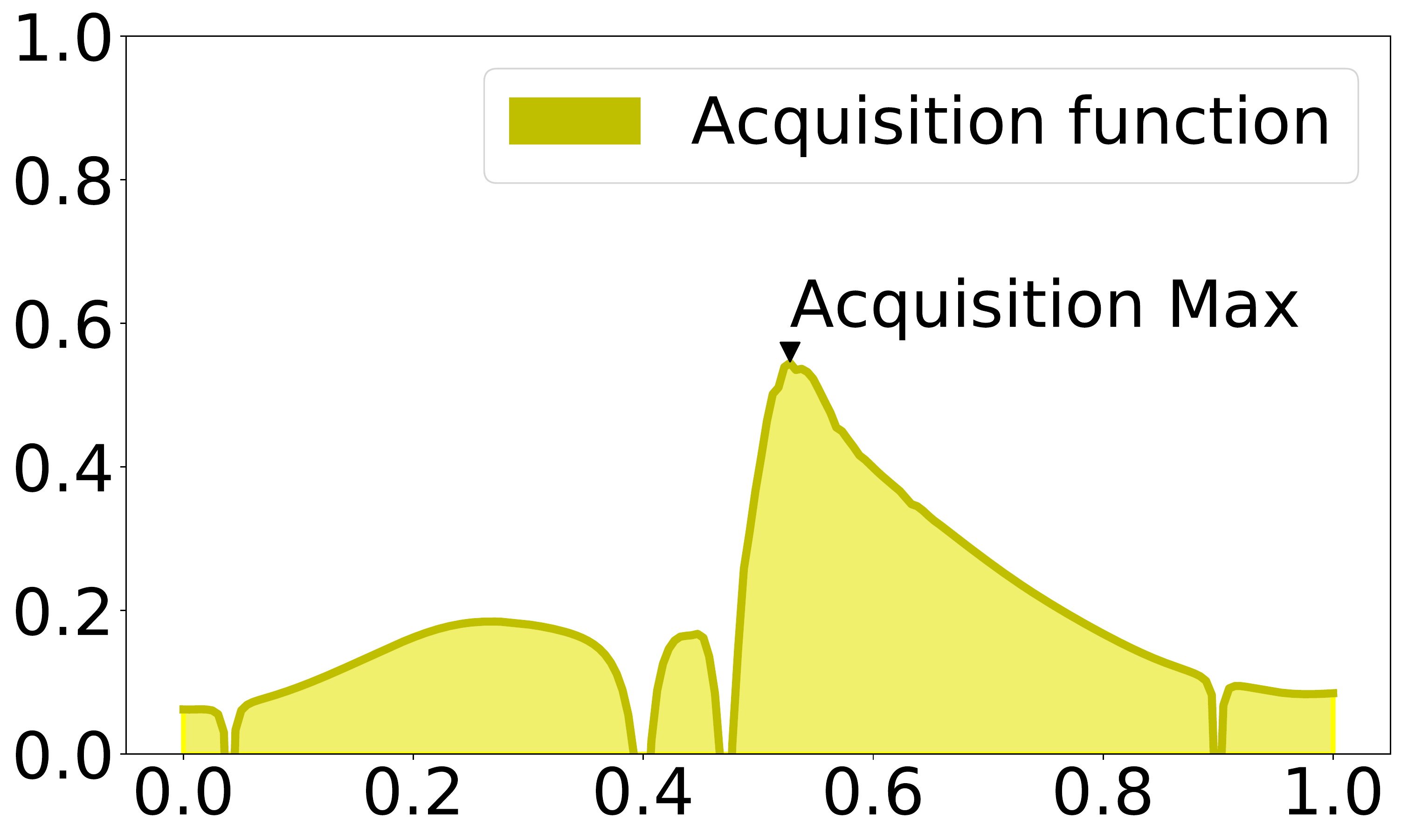} \\
\end{tabular}
\caption{\label{bo:illustration}
An example of BO on a toy 1D noiseless problem, where a function (shown in red, dotted) 
sampled from a GP prior is optimized. The top figures show the GP (mean and standard
deviation) estimation of the objective $f(\cdot)$ in blue. The acquisition
function PES is shown in the bottom figures in yellow. The acquisition is high where the
GP predicts a low objective and where the uncertainty is high. From left to right, iterations (t) of
BO are executed and the points (black) suggested by the acquisition function are
evaluated in the objective, we show in red the new point for every iteration. In a small number of iterations, the
GP is able to almost exactly approximate the objective.
}
\end{figure}

The key for BO success is that evaluating the acquisition function
is very cheap compared to the evaluation of the actual objective, because it only depends on the GP predictive
distribution for the objective at any candidate point. The GP predictive
distribution for $f(\vecb{\theta}_t)$, the candidate
location for next iteration, is given by a Gaussian distribution characterized by
a mean $\mu$ and a variance $\sigma^2$ with values
\begin{equation}\label{eq:gp}
\begin{aligned}
	\mu & = \vecb{k}_{*}^{T}
	(\mat{K}+\sigma_{n}^{2}\mat{I})^{-1}\vecb{y}\,, \\
\sigma^2 & = k(\vecb{\theta}_t,\vecb{\theta}_t) -
	\boldsymbol{k}_{*}^T(\mat{K}+\sigma_{n}^{2}\mat{I})^{-1}\vecb{k}_*\,.
\end{aligned}
\end{equation}
where $\vecb{y} = (y_1, \ldots, y_{t-1})^t$ is a vector with the objective values observed so
far; $\sigma_n^2$ is the variance of the additive Gaussian noise; $\vecb{k}_*$ is a vector with the prior covariances between
$f(\vecb{\theta}_t)$ and each $y_i$; $\mat{K}$ is a matrix with the prior
covariances among each $y_i$; and
$k(\vecb{\theta}_t,\vecb{\theta}_t)$ is the prior variance at the candidate
location $\vecb{\theta}_t$.  {The} covariance
function $k(\cdot,\cdot)$ is pre-specified; for further
details about GPs and example of covariance functions we refer the reader to
\cite{rasmussen2004gaussian}.
Four steps of the BO process are illustrated graphically in Fig.
\ref{bo:illustration} for a toy minimization problem.

In BO methods the
acquisition function balances between exploration and
exploitation in an automatic way. A typical choice for this function is the
information-theoretic method Predictive Entropy Search (PES)
\cite{hernandez2014predictive}. In PES, we are interested in maximizing
information about the location of the optimum value, $\vecb{\theta}^*$, whose
posterior distribution is $p(\vecb{\theta}^*|\mathcal{D}_{t-1})$. This can be done
through the negative differential entropy measure of
$p(\vecb{\theta}^*|\mathcal{D}_{t-1})$. Through several operations, an approximation to
PES is given by
\begin{align*}
	a(\vecb{\theta}) = H[p(y|\mathcal{D}_{t-1},\vecb{\theta})] -
	\mathbb{E}_{p(\vecb{\theta}^*|\mathcal{D}_{t-1})}[H[p(y|\mathcal{D}_{t-1},\vecb{\theta},\vecb{\theta}^*)]]\,,
\end{align*}
where $p(y|\mathcal{D}_{t-1},\vecb{\theta},\vecb{\theta}^*)$ is the posterior predictive
distribution of $y$ given $\mathcal{D}_{t-1}$ and the
minimizer $\vecb{\theta}^*$ of $f$, and $H[\cdot]$ is the differential entropy. The first
term of the previous equation can be analytically solved as it is the entropy of
the predictive distribution and the second term is approximated by Expectation
Propagation \cite{minka2001expectation}. We can see an example of the PES
acquisition function in Fig. \ref{bo:illustration}.

\section{Numerical experiments}\label{sec:exp}
Since we will consider networks of different node size $p$,
we will use in our experimental setting as the validation measure a normalized version of SHD with
respect to the maximum edge number $p(p-1)/2$. The significance level $\alpha$
will be represented for the BO algorithm as a real variable whose range lies in
the decimal logarithmic space $[-5,-1]$. The statistical test will be represented using 
a categorical variable whose value indicates one of the above mentioned four tests. 
Namely, two test based on the partial correlation coefficient: a Gaussian test based on 
the Fisher's Z transform and the Student's T test; and two test based on the mutual 
information: the $\chi^2$ test, and a test based on the shrinkage James-Stein estimator. 
As outlined before, this problem is specially suitable for BO,
since we do not have access to gradients, the objective evaluations may be 
expensive and they may be contaminated with noise.

We have employed
\texttt{Spearmint} (\url{https://github.com/HIPS/Spearmint}) for BO and
the \texttt{pc.stable} function from the \texttt{bnlearn} R package
\cite{scutari2010} for the PC algorithm execution. We have run BO with the PES
acquisition function over a set of Gaussian Bayesian networks generated
following the simulation methodology of \cite{kalisch2007}. That is, the absent
edges in the acyclic digraph $G$ are sampled by using independent Bernoulli random
variables with probability of success $d = n / (p - 1)$, where $p$ is the vertex
number of $G$ and $n$ is the average neighbor size. The probability $d$ can be
thought of as an indicator of the density of the network: smaller $d$ values mean
sparser networks. The node size $p$ is obtained from a grid of values $\{25,
50, 75, 100\}$, while the average neighbor size is $n \in \{2, 8\}$. Finally,
we consider different sample sizes $N \in \{25, 50, 75, 100\}$. Therefore, we have a
total of 32 different network learning scenarios,
that are representative of those that can be found in 
\cite{kalisch2007}. We create 40 different replicas of the experiment and report
average results across them, in order to provide more robust results. In each of these replicas, the nonzero regression
coefficients in Equation \eqref{eq:linreg} are sampled from a uniform
distribution on $[0.1, 1]$, following \cite{kalisch2007}.

For BO, we have used the PES acquisition function and 10 Monte Carlo iterations
for sampling the parameters of the GP. The acquisition function is averaged
across these 10 samples. We have used the M\'atern covariance
function for $k(\cdot, \cdot)$ (Equation \eqref{eq:gp}) and the transformation described in \cite{garrido2018dealing} so that
the GP can deal with the categorical variable (the test type). We compare 
BO with a random search (RS) strategy of the average normalized SHD
error surface and with the expert criterion (EC), taken from \cite{kalisch2007}. These authors 
recommend a value of $\alpha = 0.01$ and use the Fisher's Z partial correlation test. 
At each iteration, BO provides a candidate solution which corresponds to the best observation made so far. 
We stop the search in BO and RS after 30 evaluations of the objective.

\begin{figure}[htb!]
\centering
\begin{tabular}{c}
\includegraphics[width=0.7\textwidth]{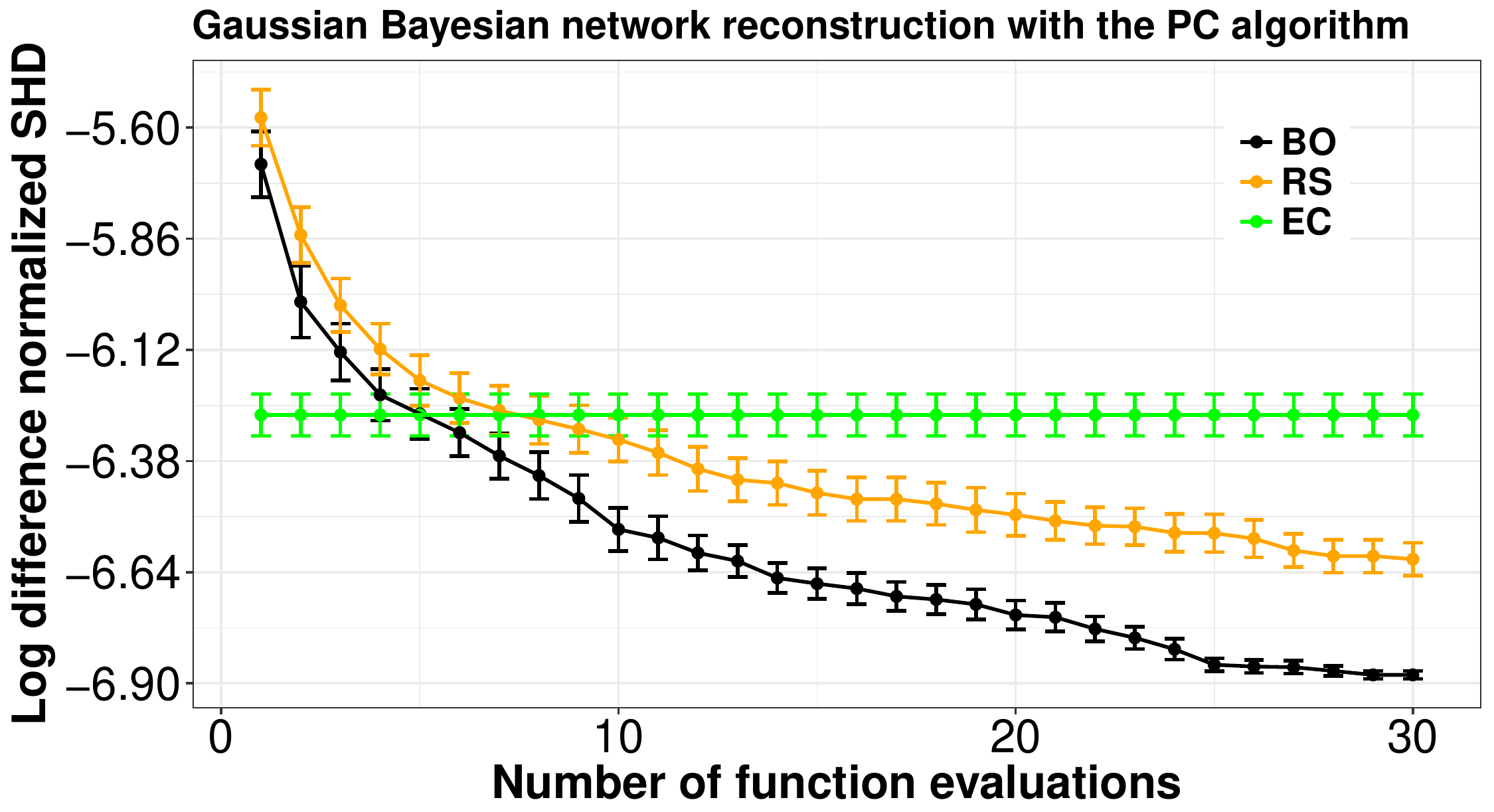}
\end{tabular}
\caption{\label{bo:plot}
Logarithmic difference with respect to the best observed average normalized SHD obtained
in 40 replicas of the 32 considered Gaussian Bayesian networks.}
\end{figure}

The average normalized SHD results obtained are shown in Fig. \ref{bo:plot}. 
We show the relative difference in log-scale with respect to the best observed result.
Therefore, the lower the values obtained, the better. We show the mean and standard
deviation of this measure along the 40 replicas of the experiment, for each of
the three methods compared (BO, RS and EC). We can see that EC is
easily improved after only 10 iterations of BO and RS. Furthermore, 
BO outperforms RS providing significantly better results as more evaluations are
performed. Importantly, the standard deviation of the results of BO are fairly
small in the last iterations. This means that BO is very robust to the 
different replicas of the experiments.

\begin{figure}[htb!]
\centering
\begin{tabular}{c}
\includegraphics[width=0.5\textwidth]{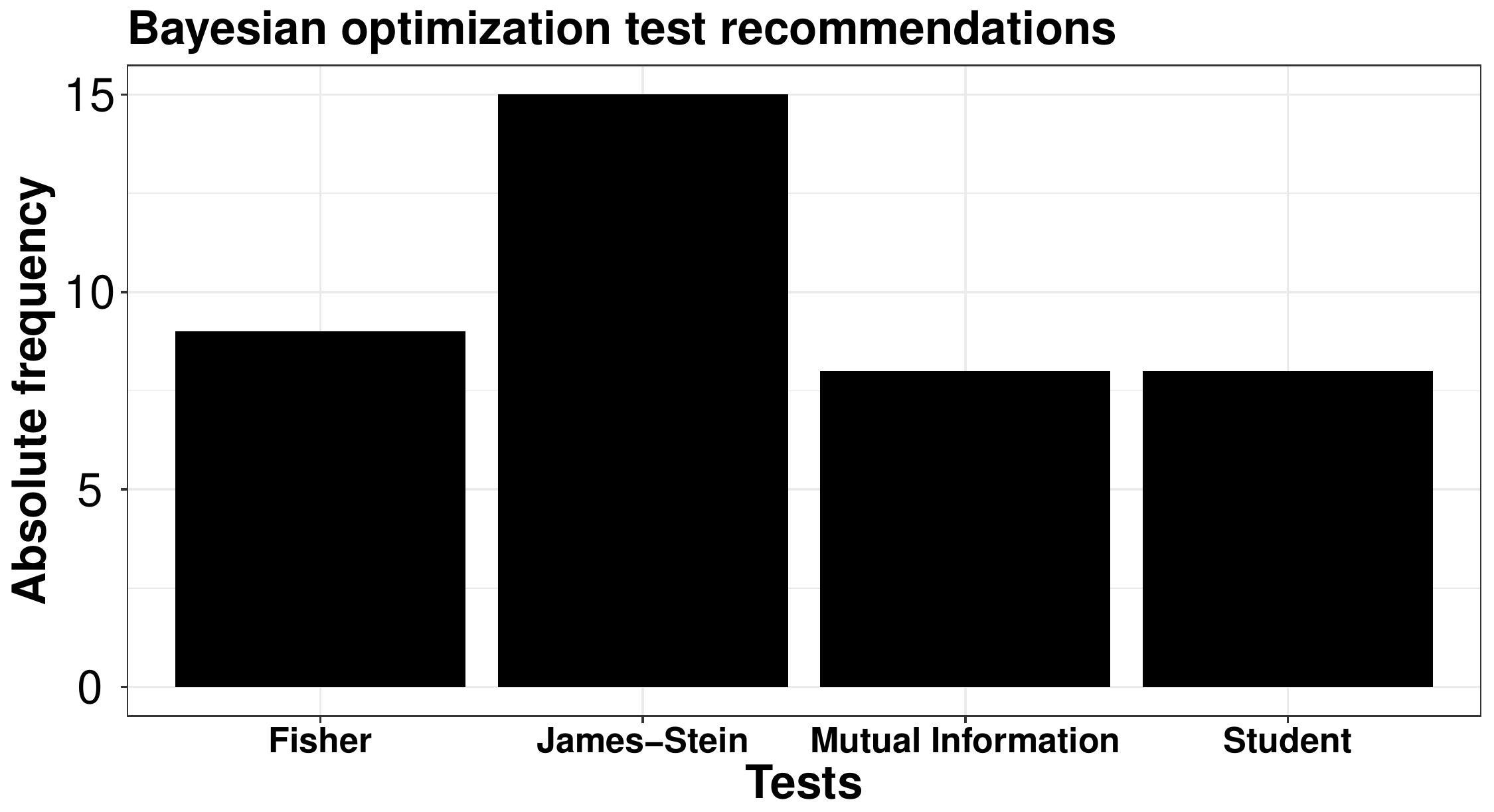}
\includegraphics[width=0.5\textwidth]{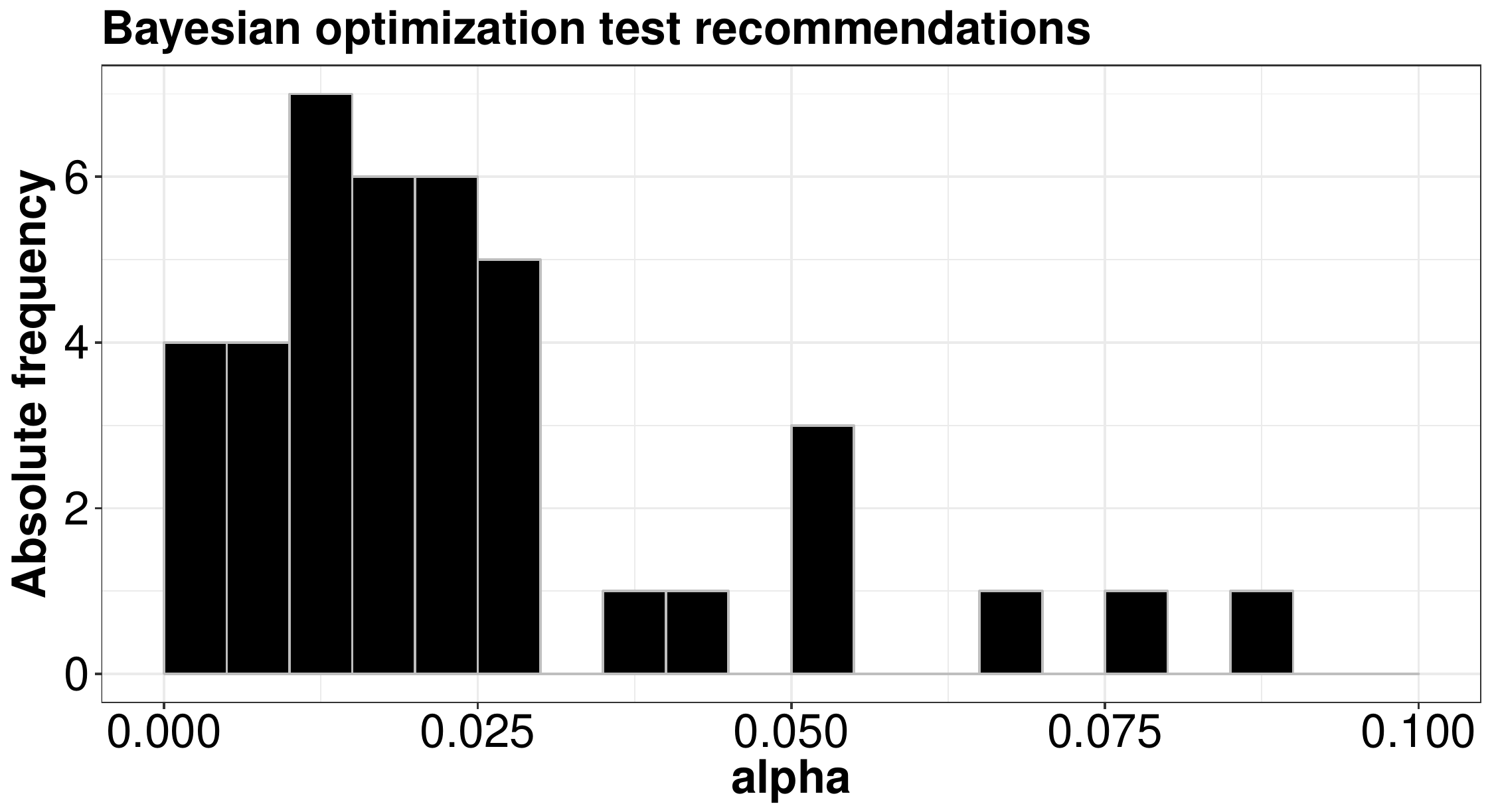}
\end{tabular}
\caption{\label{bo:hists}
Histograms with the recommended parameters by BO in the last iteration.
}
\end{figure}

Since the expert criterion is outperformed, we are interested in the parameter
suggestions delivered by BO. In order to explore these results, we have
generated two histograms that summarize the suggested parameters by BO in the
last iteration, shown in Fig. \ref{bo:hists}. We observe that the most
frequently recommended test is the James-Stein shrinkage estimator of the mutual
information \cite{hausser2009}, while the most frequent recommendation for the significance level 
is concentrated at values lower than $0.025$.

These results are very interesting from the viewpoint of graphical models
learning. The first observation is that the optimal value obtained 
for the significance level is fairly close to the one suggested in
\cite{kalisch2007}. However, the SHD results are arguably better for 
the BO than for the human expert. This may be explained by the second
interesting result we have obtained. Namely, the
shrinkage James-Stein estimator of the mutual information is suggested more times than the 
that the extended Fisher's Z partial correlation test. Therefore, in the context 
of sparse, high-dimensional networks, where we may have $p > N$ (such as in our
experimental set-up and the one in \cite{kalisch2007}), it may be 
better to focus on the selection of the statistical test, rather than 
on carefully adjusting the significance level. In the literature, however, it is 
often done the other-way-around, and more effort is put on carefully adjusting the significance level.

\section{Conclusions and future work}\label{sec:conc}
In this paper we have proposed the use of BO for selecting the optimal parameters 
of PC algorithm for structure recovery in Gaussian Bayesian networks.
We have observed that, in a small number of iterations, the expert suggestion is
outperformed by the recommendations provided by a BO method. Furthermore, an analysis of the recommendations
made by the BO algorithm shows interesting results about the relative importance 
of the selection of the statistical test, as opposed to the selection of the significance level.  
In the literature, however, it is often that the selection of the significance level receives 
more attention.

For future work, we would like to apply BO in higher dimensional settings,
where the number of nodes increases exponentially, whereas the number of samples
increases linearly. This is also a typical scenario in Gaussian Bayesian
network real applications. We would also like to explore how different error
measures, such as the true positive and false positive rates, affect the obtained results 
when they are optimized using BO. Finally, we plan to extend this methodology to consider
multi-objective optimization scenarios and also several constraints, since current BO 
methods are able to handle these problems too.

{\small 
{\bf Acknowledgements}:
We acknowledge the use of the facilities of Centro de Computaci\'on
Cient\'ifica (CCC) at Universidad Aut\'onoma de Madrid, and
financial support from Comunidad de Madrid, grant
S2013/ICE-2845; from the 
Spanish \emph{Ministerio de Econom\'ia, Industria y Competitividad}, grants TIN2016-79684-P,
TIN2016-76406-P, TEC2016-81900-REDT; from the Cajal Blue Brain
project (C080020-09, the Spanish partner of the EPFL Blue
Brain initiative); and from Fundación BBVA (Scientific Research Teams in Big Data 2016). 
Irene Córdoba is supported by grant FPU15/03797 from the
Spanish \emph{Ministerio de Educaci\'on, Cultura 
y Deporte}.
}
\bibliographystyle{splncs04}
\bibliography{paper}

\end{document}